\documentclass[preprint,12pt,authoryear]{elsarticle}
\usepackage[utf8]{inputenc}
\usepackage{graphicx}
\usepackage{pgfplots}
\pgfplotsset{compat=newest}
\usepackage{hyperref}

\journal{Computerized Medical Imaging and Graphics}

\begin{document}
	
\begin{frontmatter}

\title{Multiple Instance Learning for Digital Pathology:\\
A Review of the State-of-the-Art, Limitations \& Future Potential}
\author{Michael Gadermayr$^*$, Maximilian Tschuchnig\\
Salzburg University of Applied Sciences\\
$^*$ corresponding author, Urstein S\"ud 1, 5412 Puch/Salzburg, \\
Austria, Tel. +43 50 2211 1341, michael.gadermayr@fh-salzburg.ac.at
}

\begin{abstract}
    Digital whole slides images contain an enormous amount of information providing a strong motivation for the development of automated image analysis tools. Particularly deep neural networks show high potential with respect to various tasks in the field of digital pathology. However, a limitation is given by the fact that typical deep learning algorithms require (manual) annotations in addition to the large amounts of image data, to enable effective training. Multiple instance learning exhibits a powerful tool for training deep neural networks in a scenario without fully annotated data. These methods are particularly effective in the domain of digital pathology, due to the fact that labels for whole slide images are often captured routinely, whereas labels for patches, regions, or pixels are not. This potential resulted in a considerable number of publications, with the vast majority published in the last four years. Besides the availability of digitized data and a high motivation from the medical perspective, the availability of powerful graphics processing units exhibits an accelerator in this field. In this paper, we provide an overview of widely and effectively used concepts of (deep) multiple instance learning approaches and recent advancements. We also critically discuss remaining challenges as well as future potential.
\end{abstract}


\end{frontmatter}

\section{Motivation}

For a large range of pathologies, microscopic evaluation of biopsies is the gold standard in clinical diagnostics. Examples are smear tests, analysis of cancerous tissues during operations and postmortem histological testing. 
Due to an increasing prevalence in combination with a decrease in the number of pathologists~\citep{Barthold07a,Mudenda20a}, automated assistance tools will be of major importance in the near future. 
Digitization of slides is a first step towards efficiency enhancement which enables digital storage, easy transmission and digital processing.

Automated digital whole slide scanners are capable of digitizing complete microscopic slides within few seconds to several minutes~\citep{Hanna22a}. In an iterative process, a plurality of neighboring patches are captured with a digital camera on a high magnification level. Finally, the patches are stitched in a way that the patch boarders are indistinguishable and a single image can be accessed conveniently. Dedicated file formats~\citep{Besson19a} in combination with special image viewers allow quick and responsive visualization and interaction, in spite of the enormous size of the data.
Hardware requirements for image visualization are nowadays also modest.

Digitization alone, however, is not capable of disrupting or clearly facilitating pathologists' daily routine~\citep{Tizhoosh18a}, since digital workflows do not strongly differ from the conventional analog workflow.
In both, digital and analog workflows, effective visual examination in clinical routine requires multiple processing stages on different resolutions. Typically a screening is performed first on a low resolution, followed by a detailed view on identified relevant regions of interest.
Also, independent of digital or analogue workflows, the enormous amounts of information in combination with time-pressure in clinical routine, exhibits a potential for pathologists to miss relevant information within the data~\citep{Dano20a,VanBockstal20a,Turk19a}.

\subsection{Potential \& Limitations of Automated Image Analysis}
The automated analysis of digital whole slide images (WSIs) exhibits a high potential for a plurality of applications in the field of pathology~\citep{Tizhoosh18a}.
In recent literature, particularly segmentation~\citep{Falk2019,BenTaieb2016,Tschuchnig2022} and classification~\citep{Hou16a,Ilse19a,Lerousseau21a} tasks were considered. 
State-of-the-art methods of resolution mainly consist of deep learning models.
For both, segmentation and classification, convolutional neural networks are employed as the de facto standard method.
Segmentation can be applied as a pre-processing technique, identifying either the shape, the area and/or the number of relevant regions of interest. The output segmentation maps can be either automatically processed or used to simplify the clinician's workflow by visualizing determined relevant regions. 
Classification approaches are typically employed for means of disease type or subtype categorization. 
While classification pipelines are mostly black-boxes, segmentation output can be easily interpreted and visually validated. 
For example, while a pathologist can easily assess the quality of a segmentation map with the naked eye, it is hard or even impossible to determine the reason (or a confidence) for a certain classification output.
The categorical label also makes the integration into clinical workflows difficult. 
Even though probabilistic models~\citep{Zeinali17a} enable the computation of confidences in addition to categorical labels, a high level of transparency is thereby not achievable. 
Explainable deep learning techniques can be helpful, but straight forward approaches are not directly applicable due to the large size of WSIs~\citep{Singh20a}.
A clear limitation of segmentation algorithms is given by the fact that neural networks typically require (manually) annotated training data provided in form of a segmentation map. This, in combination with the need for sufficiently large training data, represents a clear burden for training segmentation models. 

\textcolor{black}{An additional limitation of segmentation and classification models is that they are  limited by the size of the neural networks' input images. Even with high performance computers and graphic processing units (GPUs), gigapixel WSIs cannot be processed holistically (as a whole) using modern deep convolutional neural networks.
In spite of the ongoing development and improvement of hardware and particularly graphic processing units, memory is still a limiting factor for processing digital WSIs. Although an image with a size of one gigapixel clearly fits into the memory of a single GPU, the first convolutional layer of a usual CNN, processing such an image, immediately breaks the limits of current GPUs during training.}

\subsection{Motivation on Multiple Instance Learning}
Multiple instance learning~\citep{Carbonneau18a,Ilse19a} (MIL) exhibits a category of methods partly relaxing the limitations of both, segmentation and classification. Compared to conventional classification, MIL can be applied to WSIs independently of the overall image size. Compared to the training of segmentation algorithms, there is no need to collect any local label information, e.g. by means of manually segmenting regions of interest. \textcolor{black}{Such a collection of fully annotated label maps for training segmentation models exhibits a very high manual effort (in the range of an hour per image~\citep{Joon2023,Wang2022})}. Even though ground-truth labels are available on WSI level only, MIL algorithms are partly capable of generating local predictions during the inference phase. In that sense, MIL can be interpreted as an intermediate approach between segmentation and classification. The annotations, which are typically available on WSI-level only (there is one label per WSI), can be interpreted as weak labels for local predictions. 

\subsection{Statistics}
There is a clear trend towards MIL in the field of digital pathology.
A Pubmed search based on the search-string\footnote{\href{https://pubmed.ncbi.nlm.nih.gov/?term=\%28multiple+instance+learning\%29+AND+\%28digital+pathology+OR+histology+OR+histopathology+OR+computational+pathology+OR+whole+slide+imag*\%29}{Link to Pubmed search}}\textcolor{black}{\footnote{\textcolor{black}{The asterisk in the search sub-string ("imag*") acts as a wildcard (arbitrary number of arbitrary characters) to match "imaging", "image" and "images".}}} 

\begin{table}[!ht]
\begin{tabular}{l}
\begin{tabular}[c]{@{}c@{}}"multiple instance learning" AND ("digital pathology" OR "histology"\\ OR "histopathology" OR "computational pathology" OR "whole slide imag*")\end{tabular}
\end{tabular}
\end{table}

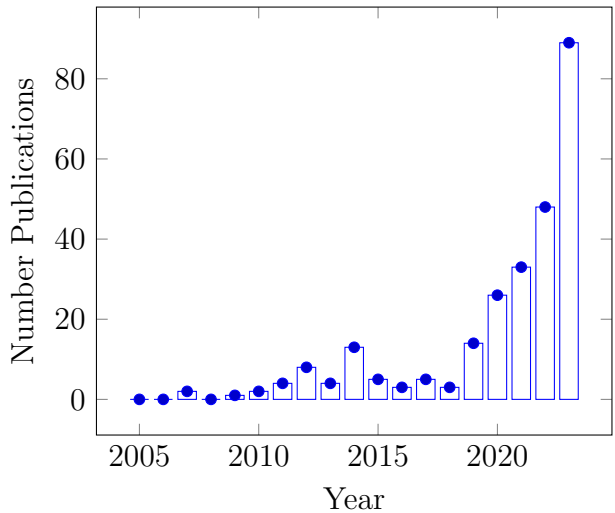
\begin{figure}[hbt]\centering
\begin{tikzpicture}
\begin{axis}[xlabel={Year},
ylabel={Number Publications}, 
ticklabel style={
    /pgf/number format/.cd,
    use comma,
    1000 sep = {}
  },]
\addplot+[ybar, bar width = 7pt] coordinates {
	(2005,0) 
	(2006,0) 
	(2007,2) 
	(2008,0) 
	(2009,1) 
	(2010,2) 
	(2011,4) 
	(2012,8) 
	(2013,4) 
	(2014,13) 
	(2015,5) 
	(2016,3) 
	(2017,5) 
	(2018,3) 
	(2019,14) 
	(2020,26) 
	(2021,33)
	(2022,48)
	(2023,89)
};
\end{axis}
\end{tikzpicture}	
\caption{Pubmed search on the combination of \textit{multiple instance learning} and \textit{digital pathology}, as well as digital pathology synonyms (date: 2023-11-13).}
\label{fig:stat}
\end{figure}

\noindent
\textcolor{black}{performed without any additional filtering delivered 260 results\footnote{\textcolor{black}{The search was performed on 15/11/2023.}}, with 75 \% (196) between 2020 and 2023 and 53 \% (137) since 2022 as shown in Fig.~\ref{fig:stat}}.
Besides established concepts, the focus of this work is on seed publications since 2020 showing novel technical approaches. \textcolor{black}{For that purpose, we mainly focus on technical contributions at the top notch conferences and journals such as NeurIPS, CVPR,  ICML, ICCV/ECCV, MICCAI, and MIA\footnote{\textcolor{black}{Abbreviations: NeurIPS: Conference on Neural Information Processing Systems; CVPR: International Conference on Computer Vision and Pattern Recognition; ICML: International Conference on Machine Learning; ICCV/ECCV: International/European Conference on Computer Vision; MICCAI: International Conference on Medical Image Computing and Computer Assisted Intervention; MIA: Medical Image Analysis}}. For application oriented work, we refer to Section 7.5.}

\subsection{Contribution}

In this paper, we provide an unstructured literature analysis of the basic building blocks of state-of-the art MIL. In addition to the basic MIL principles, we particularly focus on technical achievements developed, discussed and evaluated in recent literature since 2020. This timeframe is motivated by the steep and consistent increase in publications, shown in Fig.~\ref{fig:stat}. Based on these approaches, we provide a structured and unified mathematical and textual description and a summary of similar techniques. Finally, we provide a critical discussion about technical opportunities and limitations of current approaches, with respect to the practical application in computational pathology and hardware requirements. 

\begin{figure}[tb]
    \centering
    \includegraphics[width=\linewidth]{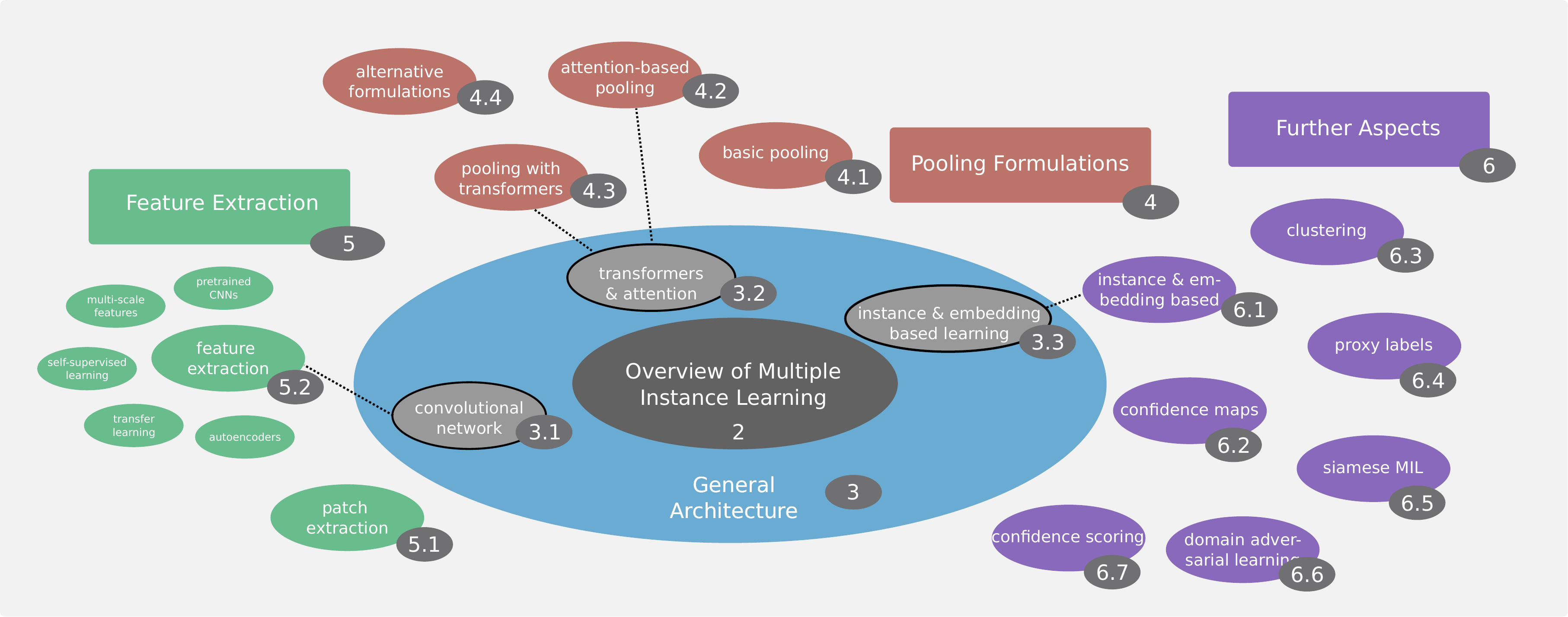}
    \caption{\textcolor{black}{Overview of the paper structure including references to sections (see bottom-right bubbles). Section 2 provides an overview, Section 3 contains the basic building blocks of general architectures and Sections 4-6 contain specific details. Connections indicate similar content. In such cases, details are provided in Section 4-6 while Section 3 provides an overview.}}
    \label{fig:differentiation}
\end{figure}

The remaining part of this paper is structured as follows. In Section~2, 
\textcolor{black}{the basic principals and an overview of MIL in digital pathology is provided. In Section~3, generic MIL architectures are provided on an abstract level representing baselines for further improvements. In Section~4, focus is on pooling for MIL aggregation. In Section~5, different methods for patch and feature extraction are discussed. Section 6 covers further remaining aspects crucial for MIL in digital pathology. A critical discussion is provided in Section~7. Section~8 concludes this paper. We decided for a structure according to technical building blocks. The publications, however, often cover several aspects. For that reason, individual papers are mentioned in several sections. Figure~\ref{fig:differentiation} provides a visual overview of this survey with links to the corresponding (sub)sections.}

\textcolor{black}{\section{Overview of Multiple Instance Learning in Digital Pathology}}

\begin{figure*}
	\includegraphics[width=\linewidth]{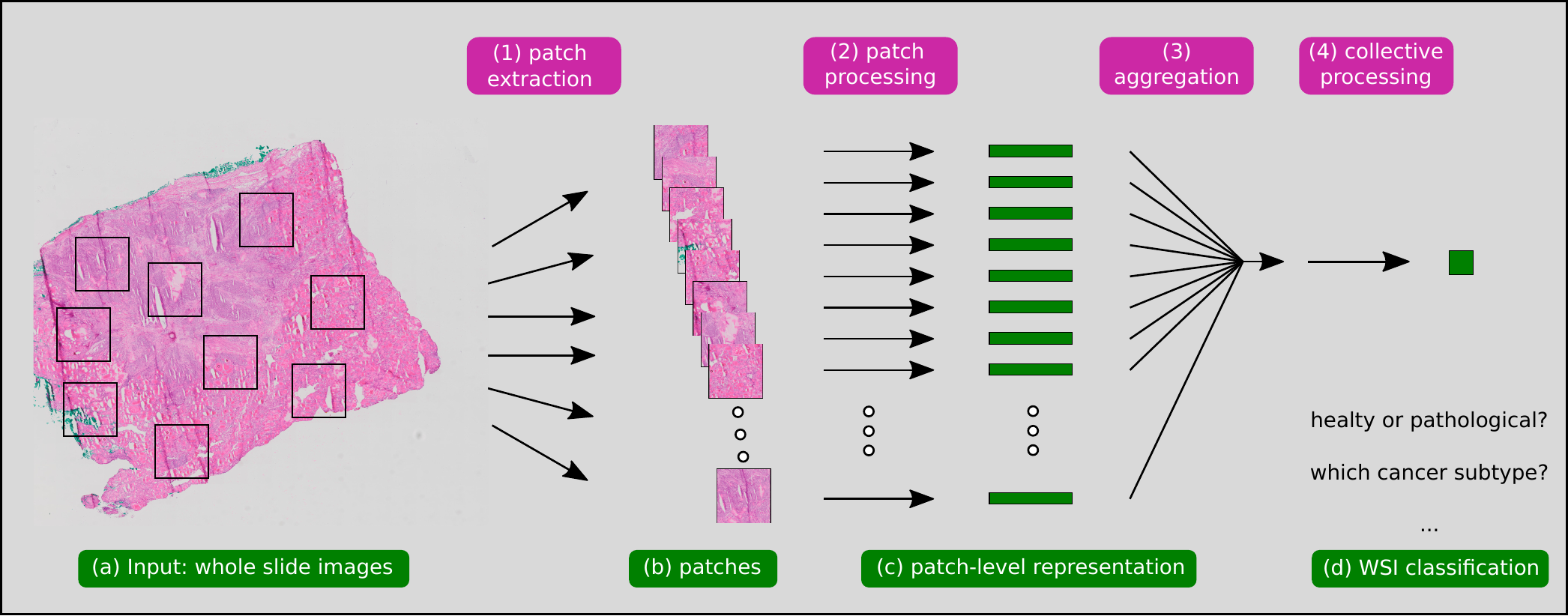}
	\caption{High-level perspective on MIL applied to WSIs. From the input images (a), patches (b) are extracted, followed by patch-level processing resulting in patch-level representations (c) and aggregation (several patch-level features to a single bag-level feature) and collective processing resulting in bag-level representations (d).}
	\label{fig:abstract}
\end{figure*}

Before the era of deep learning and deep neural networks, machine learning algorithms mostly consisted of two stages, the feature extraction stage and the classification stage. 
\textcolor{black}{While for the downstream classification task, generic algorithms were employed (such as support vector machines or random forests), feature extraction was often hand-crafted to the specific application scenario.}

\textcolor{black}{The era of deep learning changed this pipeline since deep convolutional neural networks enabled so-called end-to-end optimization of models with e.g. images as input and labels or label maps as output~\citep{He2016,Falk2019}. The feature extraction stage, consisting of (learned) image filters, is thereby integrated and automatically trained within the model.
For some applications, however, it can be advantageous to separate a trained convolutional neural network into the convolutional part, representing the feature extraction stage, and the classification part~\citep{Ianni20a}. This enables, e.g. the combination of a deep learning-based feature extraction with established classification models. 
Feature extraction models which where trained on a huge amount of data can thereby be combined with efficient classification models (with fewer parameters, such as support vector machines~\citep{Tschuchnig2022}) to achieve effective generalization in the case that a small amount of training data is available for the target application. Due to the often small number of available WSIs, this is particularly relevant for MIL.}

On a very high level, MIL approaches in digital pathology can be abstracted to the definition outlined in Fig.~\ref{fig:abstract}. 
After extracting patches (1) from the original WSIs ("bags"), each patch is first individually processed (2), followed by an aggregation (3) and a collective processing stage (4) which finally outputs a label corresponding to a "bag" of patches, which here corresponds to a WSI.

This very generic pipeline can be slightly substantiated by restricting the type of data after the patch processing stage (2).
In the case that this stage outputs a scalar (e.g. scalar between 0 and 1) for each patch, the method is referred to as \textbf{instance-based MIL} approach.
In the case that this stage outputs a feature vector for each patch, the method is referred to as \textbf{embedding-based MIL} approach~\citep{Ilse19a}.

This minor differentiation has a strong impact on the potential of the algorithms. Instance-based methods are capable of providing a final decision, individually for each patch. This output can be used to generate segmentation maps for complete WSIs indicating the relevance of different regions of interest.
This advantage directly corresponds to the disadvantage of instance-based MIL. The restriction that the information of a patch must be represented as a single scalar value potentially limits the model's power. For that reason, embedding-based MIL approaches are typically more powerful if a classification on WSI level is considered as the final goal~\citep{Butke21a,Liu2012}.

The outlined pipeline remains the same, independent whether conventional or deep learning-based models are employed. In the following, focus is on state-of-the-art deep learning architectures.

\vspace{1cm}

\section{General Architectures} 

\subsection{Convolutional Neural Networks}

\begin{figure*}
    \centering
    \includegraphics[width=\linewidth]{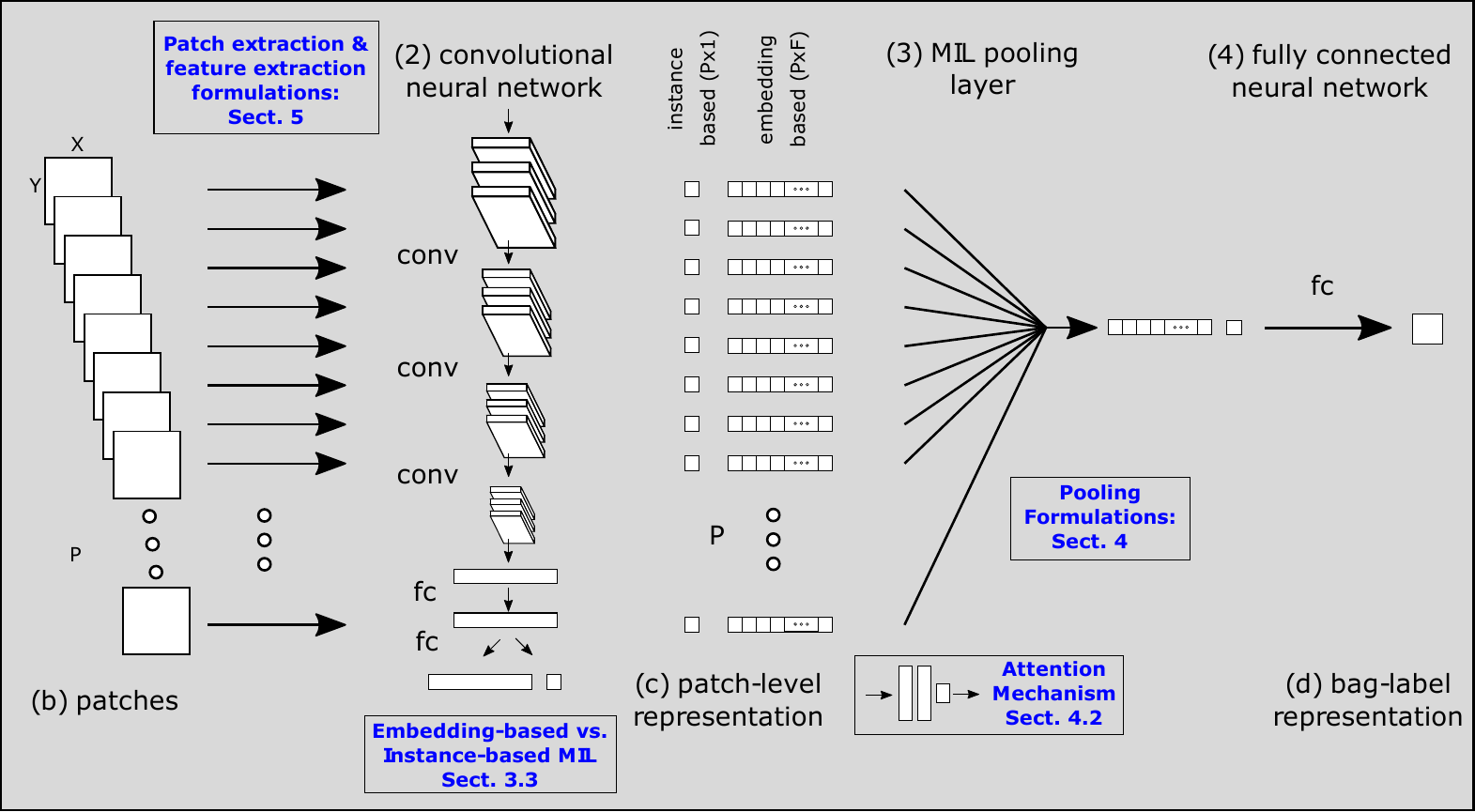}
    \caption{The scheme shown in Fig.~\ref{fig:abstract}, refined to outline typical convolutional neural network approaches and particularly the differences between instance-based and embedding-based MIL.}
    \label{fig:deepMIL}
\end{figure*}

The generic pipeline depicted in Fig.~\ref{fig:abstract} can be implemented by a deep convolutional neural network shown in Fig.~\ref{fig:deepMIL}. 

As neural network input, we consider three dimensional samples of size $P \times X \times Y$, where the constants $X$ and $Y$ refer to the patch dimension (of the extracted patches) and $P$ refers to the number of extracted patches.
While $X$ and $Y$ must be chosen to fit the characteristics of the convolutional network (described in the next paragraph), in theory, $P$ can be chosen freely. In Section~\ref{sect:discussion}, we discuss about further restrictions due to memory demands.

The first layers the patches are fed into, are convolutional layers (conv). Even though the input signal is three dimensional, only two dimensional filters are used here in a way that every single patch is processed individually (since the order in the third dimension is arbitrary and not meaningful). Often one of the well studied 2D ResNet models is used for that purpose~\citep{He2016}. As a final step, the output of the convolutional neural network is flattened for each patch individually, resulting in a matrix, with the different patches \textcolor{black}{(indicated by $p \in \{1, ..., P\}$) as rows and the features (indicated by $f \in \{1, ..., F\}$)} as columns. In instance-based MIL, $F$ is set to $1$.

This matrix is then aggregated by means of a pooling function. In theory, any differentiable function, projecting a $P \times F$ matrix ($M$)  to a vector of length $F$ is applicable here. Typical MIL Pooling functions are outlined in Section~\ref{sect:pooling}.
During this pooling operation, features per patch are converted into features per WSI. The final vector represents a descriptor for the complete histological slide.

To obtain a bag-level label in the end, either the output of the pooling function can be used (instance-based MIL) or further neural network layers are applied (embedding-based MIL). Typically for this purpose fully-connected (fc) layers are applied.

Since all operations are differentiable, this pipeline can be trained end-to-end, i.e. with pairs consisting of input patches and WSI labels. All parameters of the models can be trained at once using optimization algorithms, such as stochastic gradient descent with back propagation. To save (GPU) memory and speedup training, however, feature extraction is often performed separately.

\textcolor{black}{\subsection{Transformer Modules \& Attention}}
\textcolor{black}{
Recently transformers showed large success in the field of image analysis~\cite{Han2022,Khan2022}. In modern MIL, transformer modules are used at different stages in the architecture. Due to the huge size of the WSIs, these modules are typically not directly used for extracting features from the patches. However, transformers and other attention formulations are employed for means of pooling to aggregate patch level features~\citep{Shao21a,Li2021,Dosovitskiy2021}.
Transformers are also used as a later stage of patch level feature processing to incorporate inter-patch connections~\cite{Ren2023}.
This potentially infringes the architecture shown in Fig.~\ref{fig:abstract}. While the multi-layer perceptrons are applied on patch level ((2) patch processing), the self-attention mechanism makes use of cross-patch information by computing pair-wise scores between the features to allow adaptive weighting. For means of simplicity, we did not add this information flow in the figure. For details on attention mechanisms and transformer-based architectures used for pooling, we refer to Sections 4.2 and 4.3.}

\subsection{Deep Instance-based vs. Embedding-based Learning} \label{sect:embeddings}

\textcolor{black}{With a minor configuration, a deep learning-based MIL architecture} can be either instance-based or embedding-based (Fig.~\ref{fig:deepMIL}~(c)).
In the case that the matrix $M$ is of shape $P \times 1$ (exhibiting a column vector), a patch is represented by a single feature and the pipeline is referred to as instance-based. The single value automatically corresponds to a score or a confidence that the patch belongs a class.
This is obtained when the convolutional neural network for patch-based feature extraction has exactly one output neuron per input patch.
If $M$ is a matrix of shape $P \times F$ with $F$ greater than one, the approach is referred to as embedding-based.
\textcolor{black}{Recently, combinations of instance- and embedding-based MIL were proposed (see Section~\ref{sect:li}).}

\textcolor{black}{\section{Pooling Formulations} \label{sect:pooling}} 

We define the constants $X$ and $Y$ as the patch dimensions and $P$ as the number of patches. The number of features is defined as $F$. 
The tupel 
$$(\vec{v_1}, \vec{v_2}, ..., \vec{v_P})$$
contains for each individual patch, the corresponding feature vector. $\vec{v_p}$ contains the features for the $p$-th patch represented as a column vector. ${v_p}_f$ is the scalar value corresponding to the $f$-th feature of patch $p$.


\subsection{Basic Pooling Formulations}

Max-pooling is defined by $\vec{y}$, such that for each tupel element $\vec{v_p}$ the maximum feature of all patches is computed by

$$y_f = \max_{p \in \{1, ..., P\}}{v_p}_f, \;\;\; f \in \{1, ..., F\} \; .$$

Mean-pooling is defined such that for each feature, the arithmetic mean over the patches is computed by

$$y_f = \frac{1}{P} \cdot \sum_{p=1}^{P}{v_p}_f, \;\;\; f \in \{1, ..., F\} \; .$$




\textcolor{black}{The log-sum-exp (LSE) pooling is a continuous relaxation of the max-pooling, defined as}

$$y_f = r \cdot \log (\frac{1}{P} \cdot \sum_{p=1}^{P}r \cdot e^{{v_p}_f}), \;\;\; f \in \{1, ..., F\}$$

with r being an adjustable hyperparameter ($r > 0$)~\citep{Ilse19a}.

\subsection{Attention Mechanism} \label{sect:attention}

Conventional pooling methods suffer from the limitation that the pooling method must be chosen manually and does not contain any trainable parameters (other than LSE). Attention-based pooling~\citep{Ilse18a, Ilse19a} makes use of the idea that each feature vector is weighted using the factor $a_p$, providing a measure for the importance of the patch with respect to the final decision. Attention-based pooling is given by

$$y_f = \sum_{p=1}^{P} a_p \cdot {v_p}_f, \;\;\; f \in \{1, ..., F\} \; .$$ 

\noindent
The parameters $a_p$ are computed as follows~\citep{Ilse18a, Ilse19a}

$$a_p = \frac{e^{\vec{w}^T \cdot tanh(W_1 \cdot \vec{v_p}^T)}}{\sum_{i=1}^{P} e^{\vec{w}^T \cdot tanh(W_1 \cdot \vec{v_i}^T)}}$$

\noindent
with $\vec{w}$ being a trainable column vector of length $L$ (latent space dimensionality). The matrices $W_1$ ($L \times F$) contains trainable parameters. Since the equation for $a_p$ uses the $tanh$ sigmoidal non-linearity, with almost linear behaviour in the interval $[-1, 1]$,  a second non-linearity in the form of a sigmoid function (such as a logistic function) can be added (in combination with a second matrix $W_2$ ($L \times F$)) to the calculation of $a_p$ leading to the formulation

$$a_p = \frac{e^{\vec{w}^T \cdot ( tanh(W_1 \cdot \vec{v_p}^T) \odot sigm(W_2 \cdot \vec{v_p}^T))}}{\sum_{i=1}^{P} e^{\vec{w}^T \cdot tanh(W_1 \cdot \vec{v_i}^T) \odot sigm(W_2 \cdot \vec{v_i}^T)}} \; .$$

\noindent
$\odot$ refers to the element-wise (Hadamard) product.
To be precise, these mechanisms can be considered as self-attention mechanism, since it makes use of input from a single network layer.
\cite{Rymarczyk2021} adapted another self-attention mechanism (originally introduced by \cite{Zhang2019}) to model dependencies between instances within one bag in MIL. This is achieved by transforming the instances (referred to as $a$ and $b$) into two feature spaces $W_3 \cdot \vec{v_a}$ and $W_4 \cdot \vec{v_b}$. The matrices $W_3$ and $W_4$ are (trainable) matrices of size $\frac{F}{k} \times F$ with $k$ being a hyper-parameter to reduce the dimensionality. These feature spaces are combined using the inner product $s_{ab} = \left \langle W_3 \cdot \vec{v_a}, W_4 \cdot \vec{v_b} \right \rangle$ (to measure similarity) and are further employed to calculate $\beta_{ab}$ such that

$$\beta_{ab} = \frac{e^{s_{ab}}}{\sum_{i=1}^p e^{s_{ab}}} \; . $$

\noindent
Based on $\beta_{ab}$, for a certain instance ($a$), $o_a$ can be calculated using the following expression, with $W_5$ (which is a $\frac{F}{k} \times F$ matrix) and $W_6$ ($F \times \frac{F}{k}$ matrix) as further trainable matrices such that

$$\vec{o_a} = W_6 \cdot \sum_{p=1}^P \beta_{ap} \cdot W_5 \cdot \vec{v_p} \; ,$$

\noindent
with the sum ($\sum$) being an element-wise addition.
This leads to a mapping of the original tuple $(\vec{v_1}, \vec{v_2}, ..., \vec{v_P})$ to the transformed space $(\vec{w_1}, \vec{w_2}, ..., \vec{w_P})$ defined by

$$\vec{w_p} = \mu \cdot \vec{o_p} + \vec{v_p}$$

\noindent
using a trainable scaling parameter $\mu$. Finally, this transformed space can be used to obtain an aggregation, by utilizing the self-attention based pooling method, leading to
$$y_f = \sum_{p=1}^{P} a_p \cdot {w_p}_f, \;\;\; f \in \{1, ..., F\} \; .$$

\cite{Oner2022} developed an approach based on so-called distribution pooling. Based on the Gaussian assumption, the parameters of the normal distribution are estimated, rather than single scores, such as mean or max. Here the parameters of the marginal distributions are estimated, individually for each feature (since capturing the joint distribution is computationally infeasible~\citep{Oner2020}).

\textcolor{black}{\cite{Campanella2019} performed a large study on learning-based pooling methods, which could be interpreted as related to the attention mechanism. They investigated different classification models (such as recurrent neural networks) to directly classify the features of all patches with implicit pooling.}

\textcolor{black}{
\subsection{Pooling with Transformer Architectures}
Transformer architectures make use of the (multi-head) self-attention mechanism to encode the interactions between patches while the multi-layer perceptrons are applied, individually to each single feature vector. A transformer module includes a (multi-head) self-attention layer, a multi-layer perceptron, residual connections and layer norm.
\cite{Shao21a} proposed a transformer architecture including two transformer modules. Between two transformer layers, a position encoding strategy is performed, based on 2D convolutional layers incorporating the 2D structure of the input images.
\cite{Li2021} also proposed a transformer-based MIL approach. In contrast to \cite{Shao21a}, they used the so-called deformable attention mechanism to increase flexibility of the attention mechanism by relaxing from the restriction of a regularly sampled grid.}

\cite{Shao21a} applied transformer encoders in combination with pyramid position encoding generators in order to pool patch embeddings extracted by a ResNet. Similarly, \cite{Li2021} combined patch embeddings with positional patch encodings and a deformable transformer based encoding based pooling approach. \cite{Ren2023}~further adapted the transformer pooling idea by combining both an instance and a bag level loss function. The method consists of patch extraction using EfficientNet, mapping these features using a residual transformer backbone, and two separate streams for classifying instances and bags. 

\textcolor{black}{\subsection{Alternative Pooling Formulations}
Although deep neural networks are able to reduce the dimensionality of single patches, a large number of patches are still arising from WSIs. For dealing with such large data, structured state space models are an alternative for
sequence modelling, specifically designed for the efficient modelling of
long sequences. \cite{Fillioux2023} proposed a structured state space model invoking an optimal projection of an input sequence into memory units that compress the entire sequence, effectively pooling the sequence.}

\textcolor{black}{Another approach is to make use of graph neural networks. \cite{Zhao20a} proposed an approach based on graph convolutional neural networks based on a heuristic to construct a graph from a bag of instance features. The connectivity depends on the distance between the individual instances in the features space. Finally the bag-level class label is predicted from the graph.}

\section{Patch \& Feature Extraction}

    Here we focus on the first stage of the generic pipeline, consisting of patch extraction and feature extraction.

\subsection{Patch Extraction}

	\textcolor{black}{Patches are typically extracted at random positions within the region of interest or in a rectangular grid for} both training and testing. If a confidence map should be generated, sampling for the inference phase need to be performed in a regular grid. To increase resolution of the confidence maps, patches can be extracted with overlaps.
	The patch size typically varies between $150 \times 150$ and $224 \times 224$ pixels~\citep{Butke21a,Lerousseau21a} showing typical input sizes of two dimensional convolutional neural networks~\citep{He2016}.
	
	Since there is a correspondence between patch size and memory consumption and the number of patches and the memory consumption, both variables need to be selected with the memory restrictions in consideration. Increasing the number of patches ($P$) leads to a linear increase in the memory needed to store the feature maps of the convolutional network (Fig.~\ref{fig:deepMIL}~(2)). Increasing the patch size ($X,Y$) requires a changed architecture (particularly the fully-connected layers in (2)) with varying impact on the memory consumption. 
	
	Basic patch selection is typically performed as a preprocessing step. Usually, patches are extracted in regions showing mainly tissue and no background.
	Attention-based pooling has a similar effect, since invaluable information does not contribute to the final image representation. However, compared to a separate patch selection stage, an implicit selection in an end-to-end approach corresponds to an increased memory footprint. 
	\cite{Ianni20a} introduced patch selection based on a CNN pipeline including multiple stages, consisting of image normalization, patch selection and classification. For patch selection, the authors make use of a supervised setting by collecting segmentation masks from medical experts. The setting thereby slightly differs compared to the typical MIL setting with end-to-end training. The advantage of this approach is given by the fact that for training the MIL approach, only relevant data is used instead of randomly selected data. 
	


    \subsection{Feature Extraction}
	For feature extraction, typically convolutional neural network architectures are utilized.
	When applying these models for MIL, there exist several options. The networks can be trained from scratch (without any pre-training). Pretraining based on large datasets (such as the ImageNet dataset~\citep{Deng2009}) can be used to appropriately initialize the weights. These weights can be used directly (pretraining-only) or be further optimized on the specific training dataset (pretrained CNN). Finally, unsupervised or self-supervised approaches, like SimCLR~\citep{chen20a} can be applied.

	\textcolor{black}{\subsubsection{Pretrained CNN}}
	A generally applicable approach for feature extraction is the employment of pretrained networks (\cite{Lerousseau21a}). Training a powerful CNN, such as ResNet~\citep{He2016} with a large dataset, e.g. the ImageNet dataset, is a well studied approach in the field of image analysis~\citep{Xie2018,Frid-Adar2018} and the field of digital pathology~\citep{Tschuchnig2022}. In the latter case, training from scratch is often inhibited since patch-level annotations are not available.
	Typically, the output of a CNN's last convolutional layer is flattened and used as generic yet powerful image representation.
	The advantage of this setting is clearly the generic and efficient applicability. Dedicated training could be inhibited by the absence of (large) labeled training data, or by the absence of a powerful computing infrastructure. Pretrained CNN models can also be downloaded and immediately used.
	The disadvantage of pretrained CNNs is given by the fact that they are not optimized with respect to the application scenario. Even though generic features proved to work well in many domains, individual training can lead to improved representations, in turn leading to higher scores obtained during the downstream classification.

	\subsubsection{Transfer Learning}
 	To exploit both, powerful, large image datasets on the one hand and the incorporation of peculiarities of the specific application scenario on the other hand,  pretrained networks can be adapted as performed by \cite{Zhao20a}). 
 	 \textcolor{black}{A CNN, pretrained on a large dataset, was adapted to the specific dataset, by performing several epochs of training after initializing the weights based on the pretraining. Thereby the need for huge amounts of data can be relaxed since the feature extraction layers are initialized appropriately.} However, even though the parameters in transfer learning are well initialized it is important to keep an eye on overfitting, especially in case of small datasets~\citep{Xie2018}.

 	\subsubsection{Autoencoders}
 	Autoencoders are neural networks which are optimized to generate a latent code, based on the criterion that the output regenerates the input of the network. Since the latent code typically shows lower dimensionality, the compression during encoding is intended to maintain the important characteristics only, providing a good image representation.
 	Variational autoencoders~\citep{Wei2020} adapt this idea by making use of multivariate latent distributions instead of basic feature vectors represented in their latent code. 
 	
 	For WSI classification, \cite{Zhao20a} employed a variational autoencoder, combined with a generative adversarial model (VAE-GAN~\citep{Boesen16a}) to emphasize on generating a latent representation while enabling realistic reconstructions of the input images. 
 	To obtain a rather small, yet discriminative subset of features for each patch, the authors additionally performed feature selection to generate a compact description. To identify redundanct features, maximum mean discrepancy was applied.
 	The authors proposed an approach based on a graph-CNN to obtain decisions on a bag level, which is different from most other approaches.

	\textcolor{black}{\subsubsection{Self-Supervised Learning}}
	\cite{Li21a} made use of contrastive, self-supervised learning for feature extraction~\citep{Li21a}. Specifically, SimCLR~\citep{chen20a} was deployed for learning representations for individual patches. 
	This approach makes use of a contrastive learning strategy by training a CNN to associate the subimages from the same WSI in a set of patches.
	The model was trained to maximize the agreement between the patches from the same WSI using a contrastive loss. After CNN training, the feature extractor was used to compute the representations of the training samples for downstream tasks of embedding MIL applications. 
    
    \textcolor{black}{\cite{Liu2021} developed an approach based on a combination of contrastive learning and a conventional cross entropy loss to learn a patient-independent classifier for patches segmented from pathological images.} 
    \textcolor{black}{\cite{Liu2023} later identified an approach to deal with the challenge of class-imbalance, which is often the case in MIL. They proposed a method where
    instance features are iteratively improved using pseudo
    patch labels. This framework combines contrastive learning with a self-paced sampling scheme to ensure that pseudo labels
    are accurate.}

	\subsubsection{Multi-Scale Features}
	\cite{Li21a} proposed a pyramidal fusion mechanism to obtain multi-scale WSI features. To aggregate descriptive information from several resolutions, patch-level descriptors from two resolutions were combined by means of vector concatenation.
	In addition, \cite{Tschuchnig2020} investigated the effect of three different approaches for combining several resolution. Besides the vector concatenation approach~\citep{Li21a}, they also considered the concatenation of histograms (after the bag-of-words aggregation) and the aggregation of features from all scales into one single histogram. This approach is not based on end-to-end deep learning, but uses bag-of-words clustering and a support vector classifier.
	
	\cite{Hashimoto2020} suggested to first train individual feature extractors on single resolutions (see also Section~\ref{sect:domain_adversarial}). The individually trained feature extractors are finally combined based on an attention-pooling approach.
        \cite{Thandiackal2022} introduced a method, which learns to perform multi-level zooming in an end-to-end manner in order to save resources compared to the conventional multi-scale approach.

\textcolor{black}{\section{Further Aspects}}
\subsection{Instance and Embedding-based MIL Combinations} \label{sect:li}
Embedding-based MIL is typically more powerful, at least as it comes to the classification of complete WSIs~\citep{Butke21a}, but does not enable a scoring for single patches (finally leading to confidence maps).
\cite{Li21a} proposed a combination of instance-based and embedding-based MIL. 
The proposed dual stream MIL approach jointly trains an instance-based and an embedding-based classifier, using a dual-stream architecture. One stream uses a standard instance-based approach combined with max-pooling as show in Section~\ref{sect:embeddings} to identify the highest scoring instance. The second stream computes an attention score for each instance by measuring its distance to the so-called critical instance which is the patch showing the maximum score. To obtain a final decision on bag-level, both scores are averaged.
The attention weight in this approach is computed in a efficient way by computing the inner product between the critical instance and each individual instance. 
The goal of combining instance-based and bag-based MIL is to obtain the advantages of both approaches. 

\textcolor{black}{\cite{Wang2023} followed a similar idea to combine embedding-based and instance-based MIL by iteratively
coupling the two patchways during training.
This approach uses category information in the bag-level classifier to guide and fine-tune the (patch-level) feature extractor. 
a more accurate bag-level classifier.}
\textcolor{black}{
\cite{Ren2023} also proposed a combination of instance- and embedding-based MIL in combination with a transformer backbone. 
This method integrates transformer-based bag-level and label-disambiguation-based instance-level supervision. The latter method is based on the assignment of pseudo-labels and was motivated by reducing the impact of noisy labels.
}
\textcolor{black}{
\cite{Lin2023} proposed a method combining the two pathways in combination with interventional training. The method makes use of clustering in order to generate a dictionary of so-called confounders. 
Based on learnable projection matrices bag features and confounders  are mapped into a joint space.
}
\textcolor{black}{\cite{Gao2023} proposed a multi-task approach which combines two embedding-based streams, one for a coarse classification (pathological/healthy) and one for a fine grained classification into (cancer) subtypes. The goal of this stream separation is to train the coarse classifier without its training being influenced by the subtype classification gradient.}

\subsection{Confidence Map Generation}

In the case of instance-based MIL, the instance scores can be directly used as local confidence measures.
\cite{Li21a} proposed a hybrid approach based on a combination of both, instance and embedding-based MIL combining the advantages of both techniques.
\cite{Oner2022} suggested a pure embedding-based MIL approach, which delivers final decisions or confidences only on bag (WSI) level.
To obtain maps, first a bag of samples in a WSI within a certain region is collected and then processed similarly to complete WSIs. The predicted value for the bag is finally assigned to the center of the region. This step is repeated for each point in a rastered grid, leading to a variably resolved confidence map.

\subsection{Clustering}
\cite{Sharma21a} proposed a pipeline which makes use of local clustering as a special type of patch selection (but not for aggregation). Clustering can be employed to identify instances (patches) showing similar image features corresponding to similar image information. By performing clustering (here k-means) individually for each WSI and selecting a fixed number of instances for each cluster, the authors hypothesize that the relevant information of a WSI is more accurately approximated.
Besides the clustering approach, the authors proposed an end-to-end pipeline making use of the attention mechanism and including a loss composed of the bag loss, the instance loss (similarly to \ref{sect:li}) as well as a loss based on the Kullback-Leibler divergence~\citep{Sharma21a} between patches. The latter is applied to regularize the high instance variance of attention distribution  observed  in  similar  positive  instances. \textcolor{black}{A bag-of-visual words approach, based on clustering patch-embeddings using k-means, was proposed by~\cite{Gadermayr20a}. These clusters were further used for classification.}

\subsection{Proxy Labels}
\cite{Lerousseau21a} introduced proxy labels applied to the patches of the WSIs, depending on the patches' instance-based scores. Patches of the positive class are labeled as 
\begin{itemize}
    \item 1 such that $\alpha \%$ of the patches with the highest probability are of class 1 and
    \item 0 such that $\beta \%$ of the patches with the lowest probability are of class 0.
    \item Other patches are discarded from the loss computation.
\end{itemize}
The parameters $\alpha$ and $\beta$ are optimized during exhaustive search. 
These proxy labels can be interpreted as a quantisation of the patch level probabilities. This quantisation (or classification) enables a binary segmentation of the WSIs without requiring to choose an additional threshold value.
\textcolor{black}{A similar technique was introduced by~\cite{Su2022}. Another method, based on the attention scores was proposed by \cite{Myronenko2021}. In this approach, based on instance attention scores, three different instance-based pseudo-labels are estimated which are finally used to optimize a joint loss function.}

\subsection{Siamese MIL}
\cite{Yao2020} proposed a MIL approach making use of a Siamese network architecture and attention-based pooling. Focus here is on survival prediction which slightly differs from most other classification tasks. Dedicated to the ordinally scaled samples, this approach employs two parameter sharing convolutional networks processing samples of two different WSIs. The parallel stage is followed by the aggregation stage.
The loss is computed based on the ranking of the two inserted images. The goal is that, for pairs of subjects, the ones with the higher risk (lower survival rate) are ranked higher. The loss function contributes to the overall concordance by penalizing any discordance in any values of higher risk patients if they are greater than those of lower risk patients. Since the labels are not absolute, but allow an ordinal relation between two subjects, all other approaches for classification cannot be used for that use case.

\subsection{Domain Adversarial Learning} \label{sect:domain_adversarial}
\cite{Hashimoto2020} proposed a technique based on domain adversarial training in order to optimize the feature extraction stage in the sense that the features are invariant to variations in stain intensity. In parallel to the final layers (attention-based pooling (3) and fully connected neural network (4) in Fig.~\ref{fig:deepMIL}), a discriminator, consisting of a simple neural network, is trained to predict the domain.
In this work, each patient is treated as an individual domain so that no additional knowledge (label) on the staining condition is needed for training the feature extractor.

\subsection{Confidence Scoring} \label{sect:confidence}
\cite{Ianni20a} proposed a simple yet effective method for providing a confidence score on WSI (bag) level. For that purpose, a prediction (for each WSI) is performed multiple times. For each run, a random subset of the neurons of the neural network is omitted (here 70 \%) leading to a distribution of labels rather than a single discrete label. The distribution of predictions can be transformed into a confidence score.

\section{Discussion} \label{sect:discussion}

	Recently, a large number of approaches for classifying WSIs by means of MIL have been developed as outlined in Sections 3-6.
	
	Despite of the fact that a plurality of different medical application scenarios have been investigated in the publications, from a high level perspective the tasks are similar for most approaches.
	We identified the following goals which can be reached with MIL. The two obvious goals refer to the classification of patches (instance classification, which is a type of coarse segmentation) as well as WSIs (WSI classification). While instance classification requires an architecture which aggregates the patch information to a single value (instance-based or a combination~\ref{sect:li}), the classification of WSIs can be performed with any approach.
	Although WSI classification is an obvious goal, the hard label (numeric or binary value) thereby obtained is not optimally suited to be included into a clinical workflow. This is the case since the classification accuracy is mostly still not in a range that we can purely rely on the computer's decision. \textcolor{black}{In the case of a disaccordance additional information is needed on top of the decision.} Most models also do not even provide a confidence for a decision on the level of WSIs (see Section~\ref{sect:confidence}). Patch-based decisions are easier to interpret and can be integrated in a clinical workflow, for example to provide an estimate of relevant regions-of-interest to a pathologist. Special architectures even allow a more precise segmentation (beyond the resolution of patch sizes)~\citep{Lerousseau21a}. This allows the provision of a smooth heat map indicating relevant regions-of-interest. For integration into an effective and accepted software tool, in our opinion, this approach has \textcolor{black}{a high potential, particularly in combination with a WSI classification tool.}
    
	\subsection{Limited Hardware Ressources}
	
		In spite of the ongoing development and improvement of hardware and particularly graphic processing units, memory is still a limiting factor for processing digital WSIs.
		Even if hardware could cope with the huge images as a whole in future, it is still highly questionable if conventional convolutional architectures could be effectively trained at any point in time. The huge images would also require deeper architecture to aggregate the data and to focus on the relevant features. Also, since the number of available WSIs is often limited, we expect that training such an architectures would be extremely challenging, independently of hardware limitations. 
		
		Although MIL is capable of relaxing the challenge regarding hardware limitations, the problem is not completely solved. It is noteworthy to mention that MIL approaches (particularly if being trained end-to-end) do not analyse the whole WSI, but often only a randomly sampled subset.
		Typically, the input dimension is in the sphere of $100^3$ pixels where the first two dimensions refer to the image dimensions and the last refers to the number of patches extracted per WSI. Compared to a complete WSI (of a gigapixel), this corresponds to only one percent of the overall image data.
		Since large areas of the image are white, the patch-wise processing eliminates large unneeded information. However, in case of large tissue probes, clearly not all areas are processed and thereby considered. 
		Certain approaches focus on this challenge and propose methods for effective patch selection or sampling~\citep{Sharma21a}.
		Developments in the field of GPUs' memory will also enable a denser sampling. 
				
		We would like to further point out that there is a trade-off (with respect to memory consumption) regarding the size of the first two dimension (patch size) and the third dimension (number of involved patches). Increasing the patch size leads to a decrease in the number of patches and vice versa if the memory consumption should remain stable. We did not identify a publication explicitly focusing on the evaluation of the best trade-off. However, we also did not identify clear justifications for the chosen settings. For that reason we expect that a deep analysis could lead to further improvements.

	\subsection{Limited Training Data \& Data Augmentation}
		
		The overall amount of training WSIs are mostly limited and in the area of tens to hundreds, while the amount of pixels and extractable patches per WSI is huge.
		Since during end-to-end training, many samples correspond to a single WSI, the lack of data exhibits a clear challenge. Interestingly, data augmentation is explicitly considered in few works only. 
		
		For that purpose, \cite{Gadermayr20a} proposed a random patch sampling strategy for a conventional (not deep learning-based) approach. Based on a large amount of extracted patches per WSI, a rather small subset is randomly selected and used for training and inference. Thereby, a larger variability of samples can be created for each individual WSI. The trade-off is given by the fact that reducing the number of sampled patches per WSI reduces the quality of an individual representation, while increasing the variability. This simple approach allows performing augmentation also during inference combined with aggregation, for example using majority voting.
            \textcolor{black}{In another paper, \cite{Gadermayr23a} proposed a
        strategy to perform data augmentation in MIL settings based on the interpolation (MixUp) of patch-level features. They showed that especially a combination within WSIs can lead to improved performances.}
		\cite{Li21b} proposed a data augmentation strategy for MIL-based CT image classification (which could be also applied to digital pathology). They present a strategy to sample virtual sets of patches based on the knowledge obtained from the attention mechanism. By keeping the distribution of samples with a high attention and a low attention similar, new virtual bags are created by random sampling. Even though this method is not developed for digital pathology, it can be assumed that it is also applicable for WSIs.
        \textcolor{black}{\cite{Zhang2022} introduced internal data augmentation, by generating several virtual bags per WSI in an end-to-end pipeline. Finally two losses are used for training, one for the augmented virtual-bags and one for aggregated bags (with one bag per WSI).}
		\cite{Stegmueller22a} proposed a technique for creating virtual patches out of existing patches, by cutting and pasting small extracts. This does not solve the issue of a small number of WSIs, but can be combined with one of the approaches above~\citep{Zhang2022}. 
	
		A different approach with a similar effect as data augmentation is stain normalization~\citep{Shaban2019,Macenko2009}. \cite{Ianni20a} included image normalization by means of a convolutional neural network, which is trained end-to-end together with the classification network. In this study, focus was explicitly on real world data from multiple sites showing clear variability. \cite{Yao2020} also proposed a method including (adversarial) stain normalization in the classification pipeline.
		A standalone image normalization (or image optimization) is performed by \cite{Gadermayr20a}, by means of an unpaired generative adversarial network approach (cycle-GAN). In this method, the stain normalization is a pre-processing step not included in an end-to-end pipeline.

   	\subsection{Model Pretraining}

		Even though there exists a large amount of publications and many different datasets in the field of digital pathology, the aggregation of datasets or the use of specifically pretrained models is not observed so far. Regarding the feature extraction stage, it is common knowledge to pretrain convolutional neural networks on publicly available large generic datasets (such as the ImageNet dataset). Such techniques are also applied in the field of digital pathology. However, we did not observe publications focusing on the utilization or combination of more similar datasets showing medical or histological image content. 
		
		Firstly, digital pathology clearly distinguishes from other natural image data. Secondly, in the field of digital pathology, focus is often on similar tissue, such as cancer. Whereas organs show clearly different structure, cancer tissue shows high similarity, independently of the organ of interest. Also low-level tissue entities such as nuclei show similarities between organs.

	\subsection{Alternatives to MIL}
		Beyond classical MIL algorithms, there exist other approaches focusing on the classification of WSIs based on the same setting (with bag-labels only). 
		These methods could also be interpreted as MIL models.
		
		An example is the method proposed by \cite{Hou16a}. 
		The methods consists of a classification on patch level. The labels are aggregated into histograms and finally used for training a supervised model. Thereby the method can be interpreted as so-called count-based MIL technique.
		
		A similar method~\citep{Tschuchnig2022} uses pretrained networks for patch-wise feature extraction and performs aggregation of these features by clustering and the bag-of-words approach to obtain a feature histogram.
		Based on the histograms, finally a supervised classification model (support vector machine) is trained.

		Particularly the latter approach proved to work well with a small number of WSIs, since the feature extraction stage is based on pretrained features only. Both models use shallow classification models, such as support vector machines, which are also easy to train with few dozens of samples only.

    \subsection{Further Readings}
    \textcolor{black}{
        In this review, we identified various technical approaches for means of MIL. Due to the recently very large number of publications, we would like to state here that the review is non exhaustive. We mainly focused on the technical top-notch conferences introducing novel methods in order to identify the main ideas. We did not focus on the application side. However, since large (clinical) studies are of high importance for assessing the performance of methods, we refer to the following contributions with focus on the application: \cite{Lu2021,Campanella2019,Lu2021a,Teramoto2021,Sandarenu2022}}

        \textcolor{black}{We did not identify any similar reviews (MIL in Digital Pathology) within the last few years. In 2019, a review on weakly supervised MIL was provided by~\cite{Rony2019}. 
        \citet{Ilse19a} published a chapter on deep MIL for digital pathology including the basic concepts and a quantitative evaluation but without the latest methods. For a general overview on MIL, we refer to~\cite{Carbonneau18a,Fatima2023}. \citet{Gao2023} provide a large experimental evaluation with state-of-the art methods. \citet{Sudharshan2019} provide an evaluation including conventional (non-deep learning-based) methods. Recently, \citet{Qu2022} published a high-level survey on MIL in the field of digital pathology, providing a good literature overview but without focusing on the technical details.}

	\section{Conclusion}
	
		During the last few years, numerous approaches for a classification of WSIs by means of MIL were developed.
		In this review, we summarized frequently used deep learning architectures and focus on cutting edge literature and the containing novel aspects.
		Although the overall architecture in the majority of approaches is similar, we identified interesting and powerful specific modifications.
            \textcolor{black}{We did not identify a clear trend towards a certain architecture or methodology. However, we identified a trend towards methods using the attention mechanism and combining instance- and embedding-based learning.}
		The proposed approaches are generally applicable to different histological fields and are typically not handcrafted to any specific domain. As existing limitations, we identified the absence of sufficient training data, and the availability of graphics processing units with sufficient memory. In the latter case, ongoing developments can provide a further 
		boost by enabling the incorporation of more information from the huge histological images. To more effectively deal with small datasets, we identified research needs in the field of data augmentation as well as the use of transfer learning, based on similar, large available datasets.
	\bibliography{library.bib}
	\bibliographystyle{elsarticle-harv}

\end{document}